# DYNAMIC DOMAIN CLASSIFICATION FOR FRACTAL IMAGE COMPRESSION

K. Revathy[1] & M. Jayamohan[2]

Department of Computer Science, University of Kerala, Thiruvananthapuram, Kerala, India
[1]revathysrp@gmail.com
[2]jmohanm@gmail.com

**ABSTRACT**

*Fractal image compression is attractive except for its high encoding time requirements. The image is encoded as a set of contractive affine transformations. The image is partitioned into non-overlapping range blocks, and a best matching domain block larger than the range block is identified. There are many attempts on improving the encoding time by reducing the size of search pool for range-domain matching. But these methods are attempting to prepare a static domain pool that remains unchanged throughout the encoding process. This paper proposes dynamic preparation of separate domain pool for each range block. This will result in significant reduction in the encoding time. The domain pool for a particular range block can be decided based upon a parametric value. Here we use classification based on local fractal dimension.*

## 1. INTRODUCTION

The basic assumption of fractal compression is that some regions of the image resemble some other regions within the same image which can be regarded as possessing fractal nature. Fractals are geometric objects/shapes that possess self-similarity nature at different scales. The term fractal was introduced into the geometric world by Benoit B. Mandelbrot [1]. According to Mandelbrot, a set of mathematical equations when subjected to iterative transformations can yield a complex image and we can create many such natural-looking fractal images through such iterations.

It was Michael Barnsley [2] who introduced the idea of applying fractals in image compression technique by incorporating the theory of Iterated Function Systems (IFS) of J. Hutchinson [3]. According to Barnsley, an image can be represented as a set of mathematical equations and this idea of taking an image and express it as an IFS forms the basis of fractal image compression.

But Barnsley's IFS based compression was impractical due to its computational complexity. Advancement is made in this route by Arnaud Jacquin [4], by using Partitioned Iterated Function Systems (PIFS). Unlike Barnsley's approach of finding IFS for the entire image, Jacquin proposed to partition the image into non-overlapping blocks called range blocks and find an appropriate IFS (called as PIFS) for each range. Since then, Jacquin's approach is followed by the researchers in fractal image compression.

In PIFS compression technique, the image is partitioned into non-overlapping blocks called range. Another set of blocks larger than and similar to the range block, called domain, is also selected from the same image based on a similarity criterion. This required each range block to be compared with all possible domain blocks within the image. Such an exhaustive search adds immense operations which results in high time requirement to the encoding process. Fractal compression lacks a wide acceptance due to this high encoding time. A variety of criteria from different aspects have been proposed by researchers in fractal image compression to achieve a significant reduction in the computation time. All of them aimed to decrease the encoding time by reducing the size of the domain pool to be searched, but a standard approach is not yet defined.

Each selected domain is mapped to a range block using a set of affine transformation such as rotation, translation and resizing as well as the transformations in pixel intensity. The transformations are contractive, and, on applying a limited number of iterations the resultant image gets attracted to a fixed set. The domain information and the transformation coefficients corresponding to each range block are stored in the compressed file. In the reconstruction phase the transformations are applied on an initial image, mostly a background image, yielding parts of the original image. The initial image can be any image, and be of any size, irrespective of the original one. This resolution independence is a unique feature of fractal image compression which no other compression technique can offer. The decoding process is simple and fast compared to the encoding process.

It is still a question whether all natural images possess fractal nature or self-similarity, that the fractal compression technique depends on. But it is proved that by approximating this resemblance property fractal technique achieve high compression ratios. High compression ratios normally cause more data loss resulting in lesser image quality. However, fractal compression gives a satisfactory balance between this compression ratio and image quality compared to other image compression techniques. This makes the researchers to foresee this compression technique as a best solution in many image compression application issues. Another benefit of fractal compression is its ability to reconstruct the compressed image to any size without the loss of details.

## 2. DOMAIN CLASSIFICATION

Fractal compression is an asymmetric process in terms of algorithmic operations and executing time. Fractal encoding and decoding are different, one cannot be described as the reverse of the other. Encoding time is very high compared to the decoding time. Before we start encoding, we have to find a domain block from the same image which best matches the range block. In order to improve the quality of compression, the domain blocks are allowed to be overlapping. But, although there can be a significant gain in the quality of reconstructed image, this results in huge domain pools for each range block. For example, consider an image of size 1024x1024. Let the image be partitioned into 8x8 range blocks. There will be $2^{14}$ =16384 range blocks. Let the size of domain blocks be 16x16. (Most of the researchers use domains with a scale size double that of range block). Then for an extensive search, each range block shall be compared with 1009x1009 = 10, 18, 081 domain blocks. Thus the total number of comparisons will cumulate to around $2^{34}$. The time complexity can be estimated as $\Omega(2^n)$.

Table 1. Time requirements for range-domain matching

| Image size (n) | Number of range blocks | Number of domains | Number of comparisons |
|---|---|---|---|
| 64 | 64 | 2401 | 153664 |
| 128 | 256 | 12769 | 3268864 |
| 256 | 1024 | 58081 | 59474944 |
| 512 | 4096 | 247009 | 1011748864 |
| 1024 | 16384 | 1018081 | > 35000000000 |

Investigations have been done on classifying domains in order to reduce the search pool for best matching range-domain pair. The no-search algorithm by Furao [5] and the improvements suggested by Wang [6] are the methods which use minimum encoding time, but compromise more on image quality. Dauda et.al. [7] have attempted to reduce the domain pool based on DCT coefficients. Conci & Acquino classifies the domain pool based on the local fractal dimension of image blocks [8].

All these methods gain in reducing the pool size in one or other way, but the domains remain static, selected before the comparison starts. We propose methods for selecting the domain pool dynamically, based on the properties of the range block.

## 3. DYNAMIC DECISION METHOD

In conventional methods, the domain classification decisions are made as the first phase in encoding. Though this will help to reduce the search pool, the dynamic features of each range block selected cannot be taken into account. We propose to postpone the classification process making it part of the comparison module. Domain pool for each range block will be selected dynamically, considering the local features of range block.

The issue of deciding on which features of range block shall be considered for domain selection needs further investigations. We propose to use the local fractal dimension of image partitions to check for similarity.

### 3.1 Local fractal dimension

Barnsley presents fractal dimension (FD) as a quantity that can measure the similarity of two fractals . Fractal dimension estimates how densely a fractal occupies the metric space in which it lies.

Let A be the matrix representing the image. Let the space $R^2$ be covered by closed square boxes of side length $1/2^n$. Let N (A) denote the number of boxes which intersect A. Then the box counting theorem states that, the fractal dimension D of A can be obtained as

$$D = \lim_{n \to \infty} \frac{\log(N(A))}{\log(2^n)}$$

Different methods are available for estimating the fractal dimension of grayscale images. For color images, fractal dimension can be estimated either by converting it to grayscale or by averaging the fractal dimensions of each color place computed separately. Since natural images are not purely self-similar, if we calculate the fractal dimension of each range block separately it will be different from that of the entire image. However this value can be used to compare the complexities two image regions.

Conci A. in [8] has attempted to classify domains based on their local fractal dimension. The paper suggests separating the domains into two groups based on their fractal dimension. The FD of range block will be estimated and the search for a matching pair will be limited to the pool corresponding to the range FD. In our method, we use the differential-box counting method proposed by Sarkar and Choudhari [9] to calculate fractal dimension.

In our method, first the fractal dimension of the range blocks are calculated and listed in a linear array. Then the FD values of the overlapping domain blocks are calculated and listed in a balanced binary search tree. The policy is to search for a match in only those domains having fractal dimension close to that of the range. This can be done by fixing a 'fractal distance' value in advance. Here, the fractal distance is fixed as

$$D_f = (F_m - F_l)/3,$$

where $F_m$ is the maximum FD value in the set, and $F_l$ is the least FD in the set. This will allow us to limit the pool size to 1/3 of the entire set, in an average.

Using differential box-counting we get the FD values of images mostly in between 2.0 and 3.0. Now, if the FD of range block is $d_r$, we can confine our search to domains with FD in

$$d_r - D_f \leq d_d \leq d_r + D_f \quad , \text{ where } d_d \text{ is the FD of domain block.}$$

The domain pool is decided at run time, corresponding to the FD of the candidate range block. Since we use a height balanced binary tree, the searching and traversal can be performed in O($logn$) time. The time required to construct the tree is O (n). Thus the total overhead expense incurred in this is O (n) + O (logn).

### 3.2 Height balanced trees

The balance factor of a node in a binary tree is measured as the difference between the number of levels of its left subtree and the number of levels in its right subtree (or *vice versa*). A tree is said to be height balanced if the balance factor $b_f$ of every node in it satisfies the relation

$$-1 \leq b_f \leq 1$$

Height balanced binary trees have the advantage of having a stable time complexity for operations like traversal, insertion, deletion and search.

On insertion and deletion, the tree needs to be rotated to maintain the balance factor in the interval. However, the search operation is exactly similar to that in a binary search tree. The time complexity for a search in a height balanced binary search tree is estimated to O (n) in average case and worst case. There are variations on balanced trees such as red-black trees and splay trees, but AVL trees give the best stable time complexity in operations [10, 11].

## 4. EXPERIMENTAL RESULTS

A program which implements exhaustive search method and domain classification by fractal dimension suggested by Conci and the No-search method suggested by Furao has been tested for comparison with the proposed method. The no-search method results in minimum encoding time but with loss of image fidelity. Among the other methods this dynamic classification scheme consumes minimum encoding time without compromising quality. The results are given in Table 2 and Table 3. The reconstructed images of Lena and Highcourt are given. The original images were taken at different sizes, 512x512 and 256x256 pixels. The PSNR values obtained from the three methods for selected images are given in Table 4.

The programs are written in Scilab 5.2.0 and tested on a computer with Intel Core2 duo 2.53GHz processor with 2MB cache memory.

Table 2. Encoding time for images with 512x512 pixels

| Image | Encoding time in seconds | | | |
|---|---|---|---|---|
| | Exhaustive search | No search | Conci's | New method |
| Lena | 19.35 | 0.50 | 3.01 | 1.99 |
| Cameraman | 19.27 | 0.49 | 2.97 | 2.74 |
| Highcourt | 16.33 | 0.49 | 2.67 | 2.21 |
| Sachu | 19.76 | 0.52 | 3.54 | 2.30 |
| Flowers | 19.09 | 0.49 | 3.07 | 2.16 |
| X-mas | 18.67 | 0.49 | 2.55 | 2.01 |
| Boyandgirl | 15.34 | 0.50 | 2.86 | 2.13 |
| Street | 15.33 | 0.51 | 2.96 | 2.13 |
| Statue | 17.21 | 0.51 | 2.57 | 1.85 |
| Lord | 18.33 | 0.50 | 2.33 | 1.77 |
| Medical | 15.54 | 0.49 | 2.90 | 1.99 |

Table 3. Encoding time for images with 256x256 pixels

| Image | Encoding time for | | | |
|---|---|---|---|---|
| | Exhaustive search | No search | Conci's | New method |
| Lena | 6.77 | 0.23 | 1.07 | 0.65 |
| Cameraman | 5.86 | 0.23 | 0.87 | 0.64 |
| Highcourt | 7.32 | 0.23 | 0.79 | 0.72 |
| Sachu | 5.05 | 0.23 | 1.21 | 0.61 |
| Flowers | 5.78 | 0.23 | 0.99 | 0.49 |
| X-mas | 5.99 | 0.23 | 1.05 | 0.75 |
| Boyandgirl | 6.09 | 0.23 | 1.03 | 0.73 |
| Street | 6.12 | 0.23 | 0.91 | 0.69 |
| Statue | 5.99 | 0.23 | 0.79 | 0.72 |
| Lord | 5.75 | 0.23 | 0.81 | 0.68 |
| Medical | 5.52 | 0.23 | 0.85 | 0.75 |

Table 4. PSNR values for tested methods

| Image | PSNR | | | |
|---|---|---|---|---|
| | Exhaustive search | No search | Conci's | New method |
| Lena | 30.4675 | 25.8445 | 29.3450 | 30.9067 |
| Cameraman | 29.2497 | 25.3292 | 29.4356 | 28.7133 |
| Highcourt | 31.6988 | 24.5345 | 28.7078 | 29.3098 |
| Sachu | 30.4338 | 24.5508 | 28.4353 | 28.1233 |
| Flowers | 29.0450 | 23.9807 | 26.5467 | 27.0827 |
| X-mas | 32.4532 | 26.2034 | 31.5651 | 33.2961 |
| Boyandgirl | 32.7609 | 26.4998 | 29.7743 | 31.0340 |
| Street | 31.9008 | 25.9090 | 30.4532 | 32.5757 |
| Statue | 30.8974 | 28.3178 | 31.4545 | 33.4625 |
| Lord | 31.7778 | 25.6988 | 29.9819 | 31.7037 |
| Medical | 32.3421 | 27.5732 | 29.6511 | 33.4334 |

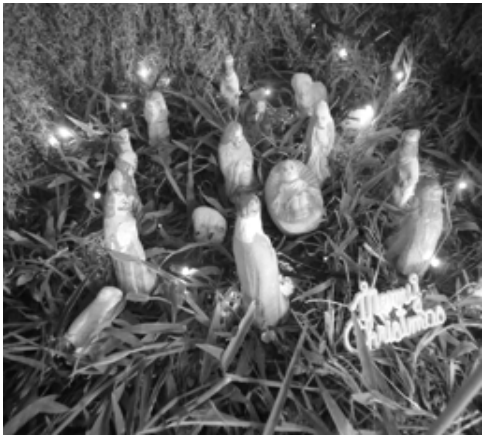 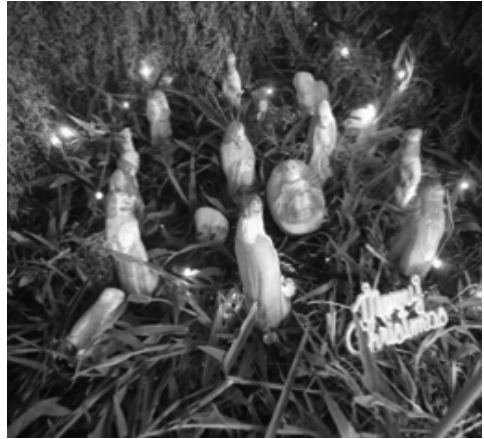

a                                                                 b

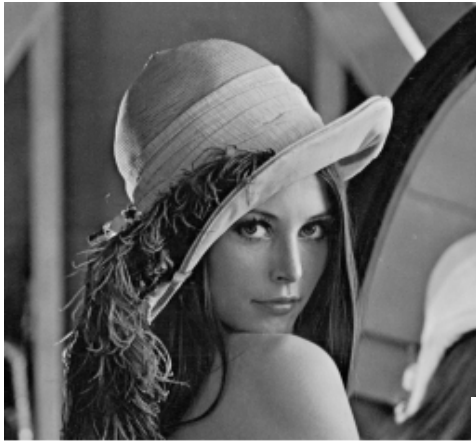 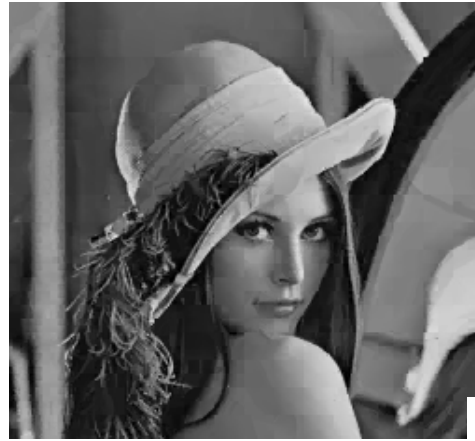

c                                                                 d

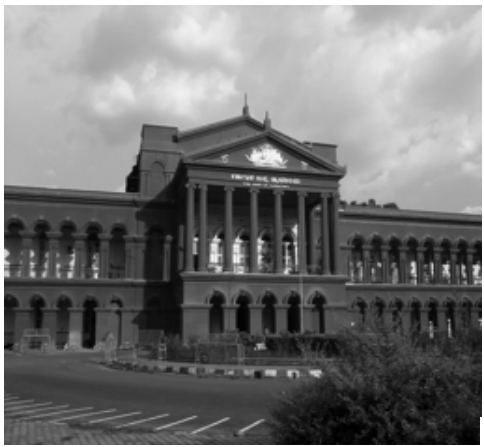 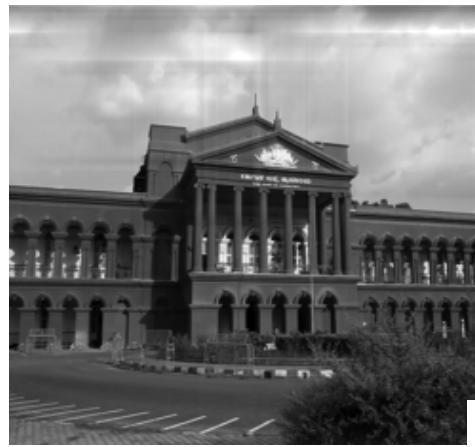

e                                                                 f

Figure 1. (a) X-mas original (b) X-mas reconstructed (c) Lena original (d) Lena reconstructed
(e) Highcourt original     (f) highcourt reconstructed

## 5. CONCLUSION

Image compression algorithms are still in search of better alternatives though there are well accepted standards. Fractal image compression technique, in spite of its unique features in resolution independence and decoding speed, is yet to find its place in the industry. We have attempted to use a dynamic decision approach on selecting the domain blocks to be compared with each range block. Experiments show that the method gains significant advantage in encoding time with good quality and compression ratio. The local features of image blocks can be used to classify domains dynamically. Further investigations are to be done on finding better similarity measures and on using better data structures.

## REFERENCES

[1] Mandelbrot, B.B. (1982) *The fractal geometry of nature,* W. H. Freeman, New York.

[2] Barnsley, M.F. (1993) *Fractals Everywhere,* Academic Press, New York.

[3] Hutchinson, John. E. (1981) "Fractals and self similarity", *Indiana Univ. Math. J.,* Vol. 30, No. 5, pp. 713–747.

[4] Jacquin, A.E. (1992) "Image Coding Based on a Fractal Theory of Iterated Contractive Image Transformations", *IEEE trans. on Image Processing*, Vol. 2, pp. 18-30.

[5] Furao S. & Hasegawa O. (2004) "A fast no search fractal image coding method", *Signal Processing and Image Communication*, Vol.19, No.5, pp. 393–404.

[6] Wang, X. & Wang, S. (2008) "An improved no-search fractal image coding method basedon a modified grey-level transform", *Computers & Graphics,* Vol. 32, pp. 445-450.

[7] Doudal, S. *et.al.* (2011) "A reduced domain pool based on DCT for a fast fractal image encoding*", Electronic Letters on Computer Vision and Image Analysis,* Vol.10, No.1, pp.11-23.

[8] Conci, A. & Aquino, F. R. (2005) "Fractal coding based on image local fractal dimension", *Computational and Applied Mathematics*, Vol. 24, pp. 83-98.

[9] Sarkar, N. & Choudhuri, B.B. (1994) "An efficient differential box counting approach to compute fractal dimension of image", *IEEE trans. on Syst. Man & Cybernetics*, Vol. 24, pp.115, 120.

[10] Tremblay, J. & Sorenson, P.G. (1991) An *introduction to data structures with applications*, Mcgraw-Hill Education, NewDelhi.

[11] Knuth, D. E. (1997) *The Art of Computer Programming,* Addison-Wesley, NewYork.

[12] Fisher, Y. (1995) *Fractal Image Compression-Theory and Application,* Springer-Verlag.

[13] Welstead, S.T. (1999) *Fractal and Wavelet Image Compression Techniques,* SPIE Press.